\documentclass{article}
\usepackage{iclr2027_conference,times}

\usepackage{amsmath,amsfonts,bm}

\def\eqref#1{equation~\ref{#1}}

\def\1{\bm{1}}

\DeclareMathAlphabet{\mathsfit}{\encodingdefault}{\sfdefault}{m}{sl}
\SetMathAlphabet{\mathsfit}{bold}{\encodingdefault}{\sfdefault}{bx}{n}

\usepackage{amsmath,amssymb,mathtools}
\usepackage{booktabs}
\usepackage{graphicx}
\usepackage{microtype}
\usepackage[table]{xcolor}
\usepackage{longtable}
\usepackage{xspace}
\usepackage{enumitem}
\usepackage{multirow}
\usepackage{placeins}
\usepackage{float}
\usepackage{fvextra}
\usepackage[most]{tcolorbox}
\usepackage{tikz}
\usetikzlibrary{arrows.meta,positioning,fit}
\usepackage{hyperref}
\usepackage{url}

\DeclareUnicodeCharacter{00B1}{\ensuremath{\pm}}
\DeclareUnicodeCharacter{00B2}{\ensuremath{^{2}}}
\DeclareUnicodeCharacter{00B3}{\ensuremath{^{3}}}
\DeclareUnicodeCharacter{00B7}{\ensuremath{\cdot}}
\DeclareUnicodeCharacter{2260}{\ensuremath{\neq}}

\definecolor{opdblue}{HTML}{2F5597}
\definecolor{opdgreen}{HTML}{3B7A57}
\definecolor{opdorange}{HTML}{C96A23}
\definecolor{draftred}{HTML}{B22222}

\newcommand{\method}{\textsc{RM-OPD}\xspace}

\newcommand{\sg}{\operatorname{sg}}

\title{RM-OPD: On-Policy Distillation for Generative Reward Models}

\newcommand{\paperauthor}[2]{\mbox{\normalfont\fontencoding{T1}\selectfont\textbf{#1}$^{#2}$}}
\newcommand{\authorline}[2]{\makebox[\dimexpr\textwidth-2\tabcolsep\relax][#1]{#2}}

\newcommand{\authorrowgap}{0.30em}
\newcommand{\authoraffiliationgap}{0.55em}
\newcommand{\authornamesep}{0.50em}
\newcommand{\affiliationsep}{0.75em}

\author{%
\authorline{c}{%
\paperauthor{Qinglin Su}{1,3},\hspace{\authornamesep}%
\paperauthor{Licheng Pan}{1,3},\hspace{\authornamesep}%
\paperauthor{Zhen Fang}{2},\hspace{\authornamesep}%
\paperauthor{Jiakang Wang}{3},\hspace{\authornamesep}%
\paperauthor{Qingnan Ren}{2}} \\[\authorrowgap]
\authorline{c}{%
\paperauthor{Yunsheng Lu}{1},\hspace{\authornamesep}%
\paperauthor{Yan Gao}{3},\hspace{\authornamesep}%
\paperauthor{Zhixuan Chu}{1,\dagger}} \\[\authoraffiliationgap]
\authorline{c}{\normalfont\fontencoding{T1}\small $^{1}$Zhejiang University\hspace{\affiliationsep}$^{2}$University of Science and Technology of China\hspace{\affiliationsep}$^{3}$Xiaohongshu Inc.}%
}

\hypersetup{
  hidelinks,
  pdftitle={RM-OPD: On-Policy Distillation for Generative Reward Models},
  pdfauthor={Qinglin Su, Licheng Pan, Zhen Fang, Jiakang Wang, Qingnan Ren, Yunsheng Lu, Yan Gao, Zhixuan Chu}
}

\iclrfinalcopy

\begin{document}

\maketitle
\begin{NoHyper}
\begingroup
\renewcommand{\thefootnote}{\fnsymbol{footnote}}
\footnotetext[2]{Corresponding authors.}
\endgroup
\end{NoHyper}
\lhead{}

\begin{abstract}
Reinforcement Learning on Mixed-domain preference data (Mix-RL) has become the mainstream approach for training general-purpose Generative Reward Models (GRMs).
However, Mix-RL suffers from two critical bottlenecks:the reward sparsity induced by outcome rewards, and the cross-domain interference arising from jointly optimizing heterogeneous capabilities, which together give rise to imbalanced capability acquisition and pervasive reward hacking.
Inspired by the success of On-Policy Distillation (OPD) in multi-domain capability integration, we introduce \textbf{\method (Reward Modeling via On-Policy Distillation)}\ , a two-stage framework that separates specialist acquisition from capability integration.
First, we cultivate domain-specialized teacher models via single-domain GRPO fine-tuning,
allowing each expert to reach its performance ceiling in isolation; Then we consolidate these teachers into one student model through (i) \textbf{routed on-policy distillation}, which provides dense signed token-level supervision on student-generated critiques;
(ii) \textbf{domain contribution reweighting}, which reweights token guidance toward target domain shares, preventing specific domains from dominating optimization; and (iii) optional \textbf{criteria context distillation}, which conditions teachers on domain-specific evaluation criteria to sharpen token-level supervision.
Across multiple RM benchmarks and two base models, \method consistently outperforms vanilla Mix-RL and other strong baselines, achieving overall superior performance and exhibiting an
emergent "teacher-surpassing" effect.
These results establish RM-OPD as a promising paradigm for building generalist GRMs. 
\end{abstract}

\section{Introduction}
\label{sec:introduction}

Generative reward models (GRMs) have emerged as a powerful approach to enhance the general capabilities of large language models (LLMs) by providing robust and more reliable reward signals~\citep{zhang2025generativeverifiers,mahan2024generativerm,chen2025rmr1}. Unlike traditional scalar reward models that compress evaluation into a numerical score, GRMs generate natural-language rationales with detailed critiques before issuing a final judgment, leading to significantly higher robustness, interpretability, and generalizability through explicit reasoning~\citep{zachary2024cloud, zhang2025generativeverifiers,yu2025criticrm}. These advantages become particularly important in general-domain reward modeling, as the evaluation criteria for rewards are more diverse and complex, and explicit references or ground-truth are often unavailable~\citep{liu2025deepseekgrm}. Generalist reward modeling is thus crucial for improving the performance of LLMs across broader applications. However, building such a generalist GRM requires distinct evaluation capabilities demanded by different domains: mathematics requires step-level logical verification, code requires reasoning about functional correctness and edge cases, and safety requires intent-sensitive judgment and calibrated refusal~\citep{lambert2025rewardbench,liu2025rmbench,zhou2025rmb}. The central challenge is therefore to consolidate these heterogeneous evaluation capabilities into a single GRM while enabling them to coexist without mutual degradation.

\begin{figure*}[t]
  \centering
  \includegraphics[width=0.98\textwidth]{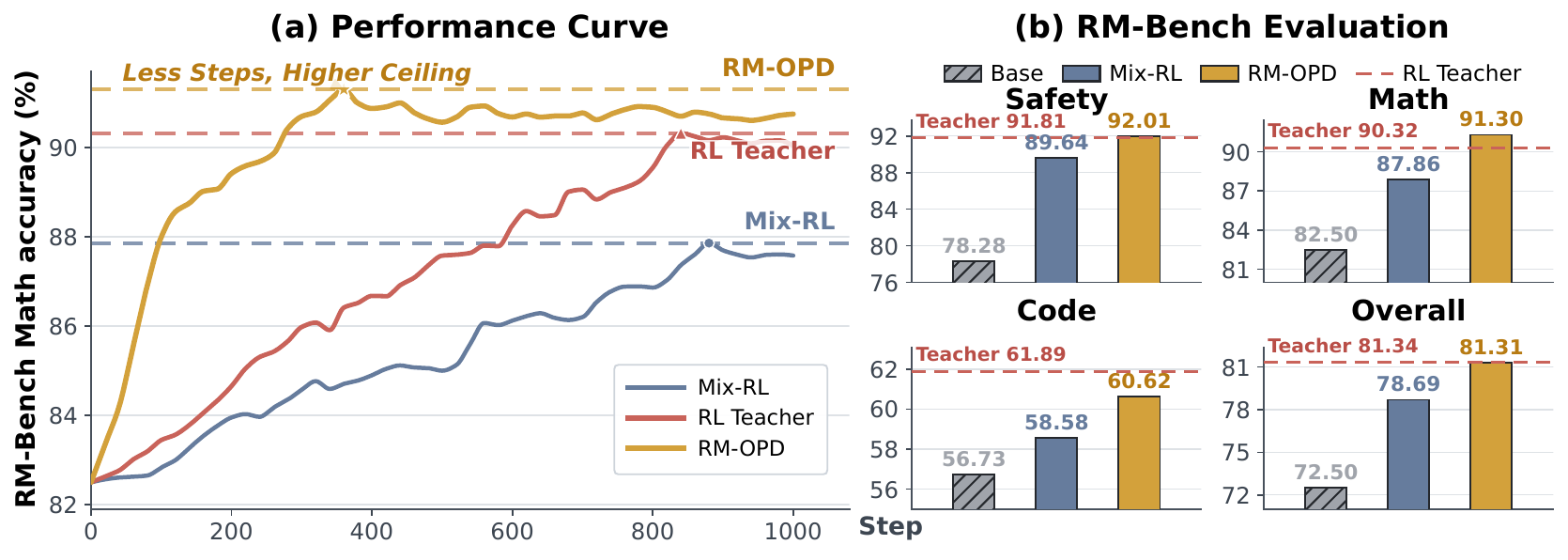}
  \vspace{-10pt}
  \caption{\textbf{Performance Comparison in Multi-domain Training.}
\textbf{(a)} On RM-Bench Math, RM-OPD converges faster than both the RL teacher and Mix-RL. In contrast, Mix-RL converges more slowly and saturates at a noticeably lower performance level.
\textbf{(b)} Across Safety, Math, and Code, RM-OPD consistently outperforms Mix-RL while closely matching or even exceeding the corresponding specialist teachers, demonstrating that RM-OPD effectively consolidates domain-specific capabilities into a single model without the performance trade-offs of Mix-RL.}
  \label{fig:teaser}
  \vspace{-15pt}
\end{figure*}

A prevailing solution for building generalist GRMs is reinforcement learning on mixed multi-domain data (Mix-RL), typically Group Relative Policy Optimization (GRPO)~\citep{shao2024deepseekmath, chen2025rmr1,liu2025deepseekgrm}, which pools prompts from various domains and optimizes one shared policy against a joint objective. Although Mix-RL improves overall performance of GRMs in multi-domain scenarios, they still fail to reach the performance ceiling of single-domain RL. As different tasks demand heterogeneous evaluation capabilities, sparse outcome rewards lack the granularity to harmonize these cross-domain interference, inducing a "seesaw effect" where optimizing specific reasoning ability (e.g., safety guard) inevitably degrades helpful reasoning like logical verification, limiting the coexistence of multi-domain capabilities~\citep{standley2020seesaw,yu2020pcgrad,lightman2024lets,wang2026rationale,ma2026mopd}. Figure~\ref{fig:teaser} demonstrates the gap between Mix-RL and independently trained specialists. This dilemma necessitates a shift to dense, trajectory-level reward signals to provide decoupled domain-specific supervision.

Recent successes of on-policy distillation (OPD) in multi-domain LLM post-training suggest a
different paradigm. OPD consolidates heterogeneous capabilities from independently trained specialists into one student model through dense token-level supervision on the student's own rollouts, which is widely adopted in frontier models~\citep{agarwal2024onpolicy,gu2024minillm,ma2026mopd,deepseekai2026deepseekv4,xiao2026mimo,zeng2026glm}. On-policy context distillation further shows that a context-conditioned teacher can supervise the student's own trajectories, enabling explicit contextual knowledge or behavioral criteria to be internalized in the student's parameters ~\citep{ye2026onpolicycontextdistillationlanguage,siyan2026opsd,jonas2026sdpo}. Together, these developments raise a pivotal question for GRMs: \emph{Can OPD be adapted to consolidate domain-specialized reward teachers into one generalist GRM while attaining the comparable multi-domain evaluation capability?}

To address this question, we introduce \textbf{\method} (Reward Modeling via On-Policy Distillation), a two-stage framework that separates specialist acquisition from capability integration. In the first stage, independent single-domain GRPO training produces domain-specialized reward teachers, allowing each capability to develop without cross-domain interference. In the second stage, \method consolidates these teachers into one student model through three coordinated components: (1) \textbf{routed on-policy distillation}, which sends each student-generated critique to its matched teacher and converts teacher--student log-probability differences into dense signed token-level supervision; (2) \textbf{domain contribution reweighting (DCR)}, which aligns each domain's update share with its target allocation and prevents any domain from dominating optimization; and (3) optional \textbf{criteria context distillation (CCD)}, which conditions teachers on domain-specific evaluation criteria to sharpen token-level supervision. Together, these components replace joint sparse-reward optimization with routed dense supervision that integrates diverse evaluation capabilities into one generalist GRM.

Figure~\ref{fig:teaser} summarizes the faster convergence and stronger capability coexistence achieved by \method. Across eight benchmarks and two base families, \method consistently improves the macro average over matched Mix-RL by by 4.60--5.55\% relative. Our contributions are threefold:
\begin{itemize}[leftmargin=*,itemsep=2pt,topsep=3pt]
  \item \textbf{Limitation analysis of Mixed-domain RL.} We identify two structural limitations of Mix-RL for building generalist GRMs as cross-domain interference and sparse outcome rewards---and show that Mix-RL leaves a consistent performance gap to independently trained specialists.
  \item \textbf{The RM-OPD framework.} We introduce RM-OPD, a two-stage framework that combines task-routed on-policy distillation, domain contribution reweighting, and domain-specific criteria context distillation to integrate specialized evaluation capabilities into one consolidated GRM.
  \item \textbf{Strong Performance and Generalization.} Through extensive experiments across multiple RM benchmarks and two base models, we demonstrate that RM-OPD achieves consistent improvements over vanilla Mix-RL and even surpasses specialized teachers in some domain, exhibiting higher multi-domain performance ceiling and robust out-of-domain generalization capabilities. 
\end{itemize}

\section{Related Work}
\label{sec:related}

\paragraph{Reinforcement Learning for Generative Reward Models.}
Generative reward models (GenRMs) cast preference evaluation as autoregressive next-token prediction, producing rubrics, critiques, and verdicts in natural language~\citep{ziyi2024beyondsrm,zhang2025generativeverifiers,mahan2024generativerm,zheng2023judging, zhuohao2025rewardanything}. To improve judgment quality beyond supervised fine-tuning, recent work applies reinforcement learning with verifiable rewards (RLVR) to optimize GenRMs against verdict correctness, exemplified by GRPO on chain-of-rubrics judgments~\citep{chen2025rmr1}, rule-based RL with inference-time voting~\citep{liu2025deepseekgrm}, and rationale-augmented variants~\citep{wang2026rmnlhf,wang2026rationale,yu2025criticrm}. However, such RLVR-based training relies on sparse verdict-only rewards and a single generalist judge, which struggle to supervise intermediate reasoning tokens and to reconcile heterogeneous domain criteria~\citep{wang2026rmnlhf,wang2026rationale}. \method instead delivers dense token-level supervision from domain-specialized experts on the student's own judgment rollouts.

\paragraph{On-Policy Distillation.}
Unlike classical offline distillation that suffers from distribution shifts and exposure bias ~\citep{kim2016seqkd}, on-policy distillation (OPD) optimizes the student over its own generated rollouts, guided by teacher-scored log-probabilities ~\citep{yaxuan2026rethinkingopd}. This paradigm has proven effective in the LLM domain through reverse-KL and policy-gradient formulations~\citep{agarwal2024onpolicy,gu2024minillm}, KL-constrained RL unification~\citep{yang2026gopd,chenxu2026rlsd}, and efficiency-oriented variants~\citep{kang2026asyncopd}. The paradigm has since been extended to multi-teacher routing~\citep{ma2026mopd,huanang2026openmopd}, flow-matching generators~\citep{fang2026flowopd}, speculative drafters~\citep{lei2026draftopd}, and self-evaluation~\citep{zhang2026see}. Despite this progress, to the best of our knowledge, no prior work has systematically applied OPD to reward modeling. \method fills this gap as the first multi-teacher OPD framework for generative reward models, consolidating domain-specialized experts into a single generalist GRM.

\section{Preliminaries}
\label{sec:preliminaries}

\paragraph{Generative reward models.}
Generative Reward Models (GRMs) formulate reward modeling as a generation task, producing preference judgments and reasoning rationales. In pairwise generative reward modeling~\citep{chen2025rmr1}, each input $z=(x,y_a,y_b)$ contains a user request $x$ and two candidate responses, with a ground-truth label $\ell\in\{a,b\}$ indicating the preferred response. A generative reward model $r_\theta$ produces a textual judgment:
\begin{equation}
  j=(j_1,\ldots,j_T)=j^{\mathrm{reason}}\oplus j^{\mathrm{verdict}},
  \qquad
  r_\theta(j\mid z)=\prod_{t=1}^{T}r_\theta(j_t\mid z,j_{<t}),
  \label{eq:genrm}
\end{equation}
where $T$ is the judgment length, $j_{<t}$ denotes preceding tokens, and $\oplus$ denotes concatenation. The reasoning segment $j^{\mathrm{reason}}$ contains a rubric, solution check, or critique, followed by a final verdict $j^{\mathrm{verdict}}$ specifying the predicted preference $\hat\ell\in\{a,b\}$. Judgment accuracy measures agreement between $\hat\ell$ and $\ell$. Within this general formulation, we focus on $\mathcal D=\{\text{math},\text{code},\text{safety}\}$, each demanding distinct evaluation criteria and judgment skills.

\paragraph{On-policy distillation.}
Knowledge distillation aims to compress teacher capabilities into a student model by minimizing their output divergence. To mitigate distribution shift, on-policy distillation (OPD) performs this alignment on student-generated trajectories, obtaining teacher feedback at prefixes encountered during autoregressive generation~\citep{agarwal2024onpolicy}. For generative reward models, the student $r_\theta$ generates $j\sim r_\theta(\cdot\mid z)$, while a frozen teacher $r_T$ evaluates the same judgment prefixes. Assuming a shared token vocabulary, a reverse-KL formulation~\citep{gu2024minillm} defines the local discrepancy at each visited state $(z,j_{<t})$ as
\begin{equation}
  \mathcal K_t(\theta;z,j_{<t})
  =
  D_{\mathrm{KL}}\!\left(
  r_\theta(\cdot\mid z,j_{<t})
  \Vert
  r_T(\cdot\mid z,j_{<t})
  \right).
  \label{eq:opd}
\end{equation}
The global objective minimizes the expectation of these local discrepancies over the student's active generation distribution. By aligning the model on its own generated trajectories, OPD effectively mitigates exposure bias and ensures robust generalization in interactive or iterative generation tasks.

\section{Motivation}
\label{sec:motivation}

\subsection{Why does Mix-RL leave capability gaps?}
\label{sec:motivation-mixrl}

RL algorithms, typically GRPO\citep{shao2024deepseekmath}, enable LLMs to learn domain-specific judgment skills through online exploration under a consistent reward. Trained in isolation, the mathematics, code, and safety teachers each attain a distinct per-domain performance ceiling, establishing the capability yardstick on their matched RM-Bench domains (Figure~\ref{fig:teaser}). This raises a natural question: \emph{can GRPO reach these per-domain ceilings simultaneously through mixed-domain training (Mix-RL)?}

Mix-RL minimizes $\mathcal L_{\mathrm{mix}}=\sum_{d\in\mathcal D}\lambda_d\mathcal L_d$ through shared parameters, where $\lambda_d$ is the weight of domain $d$. Let $g_d=\nabla_\theta\mathcal L_d$. The first-order change of domain $e$ after one gradient step is
\begin{equation}
  \Delta\mathcal L_e\approx-\eta\!\left[
  \lambda_e\lVert g_e\rVert^2+\sum_{d\neq e}\lambda_d\langle g_e,g_d\rangle\right].
  \label{eq:mix-interference}
\end{equation}
The first term gives within-domain descent, whereas the cross-domain terms characterize how updates from other domains affect $\mathcal L_e$. When $\langle g_e,g_d\rangle<0$, optimizing domain $d$ counteracts progress on domain $e$; if the accumulated interference exceeds the within-domain term, the mixed update can even increase $\mathcal L_e$ while improving the joint objective~\citep{standley2020seesaw,yu2020pcgrad}. Repeated across domains, these antagonistic updates erode previously acquired capabilities and produce the seesaw effect observed in Table~\ref{tab:motivation-mixing}.

Besides, sparse outcome rewards provide insufficient granularity to mitigate cross-domain interference. For a judgment $j=j^{\mathrm{reason}}\oplus j^{\mathrm{verdict}}$, the reward $R(z,j)=\mathbb 1[\hat\ell=\ell]$ depends only on the final verdict. The same trajectory-level advantage is therefore propagated across the entire critique without identifying which reasoning steps caused success or failure.

\noindent
\begin{minipage}[t]{0.49\textwidth}
\vspace{-5pt}
Critiques with different reasoning quality remain indistinguishable whenever they reach the same verdict, leaving the domain-specific judgment process without direct supervision. Table~\ref{tab:motivation-mixing} empirically corroborates these cross-domain trade-offs. Together, cross-domain interference and sparse outcome supervision leave Mix-RL with consistent capability gaps relative to isolated training.
\end{minipage}
\hfill
\begin{minipage}[t]{0.49\textwidth}
\vspace{-15pt}
\centering
\makeatletter\def\@captype{table}\makeatother
\caption{\textbf{Mix-RL trade-offs.} \mbox{RewardBench} accuracy (\%). \textcolor{draftred}{Red} values denote drops. Base model is DeepSeek-R1-Distill-Qwen-7B.}
\vspace{5pt}
\label{tab:motivation-mixing}
\small
\setlength{\tabcolsep}{1.3pt}
\renewcommand{\arraystretch}{1.05}
\def\drop#1{\hspace{1pt}\textcolor{draftred}{\footnotesize($\downarrow$#1)}}
\begin{tabular}{@{}l r@{\hspace{1.5pt}}l r@{\hspace{1.5pt}}l r@{\hspace{1.5pt}}l@{}}
  \toprule
  \textbf{Model}
  & \multicolumn{2}{c}{\textbf{Safety}}
  & \multicolumn{2}{c}{\textbf{Math}}
  & \multicolumn{2}{c}{\textbf{Code}} \\
  \midrule
  Base      & 71.89 &             & 86.58 &             & 86.79 & \\
  $+$Math   & 68.92 & \drop{2.97} & 96.42 &             & 91.57 & \\
  $+$Code   & 67.97 & \drop{0.95} & 93.96 & \drop{2.46} & 93.39 & \\
  $+$Safety & 82.16 &             & 92.62 & \drop{1.34} & 93.80 & \\
  \bottomrule
\end{tabular}
\end{minipage}

\subsection{How to raise the coexistence ceiling?}
\label{sec:motivation-raising}

The analysis above identifies the supervision needed to raise the coexistence ceiling. Raising the coexistence ceiling requires supervision that is \emph{on-policy}, covering states induced by the current student; \emph{domain-specific}, preserving coherent evaluation criteria; and \emph{process-dense}, reaching the critique rather than only the final verdict.

OPD matches these requirements by sampling trajectories from the current student and evaluating teacher distributions at the visited prefixes~\citep{agarwal2024onpolicy,gu2024minillm}. It preserves on-policy exploration while replacing outcome rewards with token-level guidance over the judgment trajectory. Independent domain experts can therefore provide coherent supervision without directly mixing their sparse rewards.

This correspondence raises our central question: \emph{can OPD consolidate the distinct judgment capabilities of domain-specialized reward teachers into one generalist GRM?} We answer this question with \method (Reward Modeling via On-Policy Distillation), a two-stage framework that separates domain expertise acquisition from capability integration.

\FloatBarrier

\section{\method}
\label{sec:method}

\begin{figure}[H]
  \centering
  \includegraphics[width=1.00\textwidth, trim=3pt 50pt 3pt 45pt, clip]{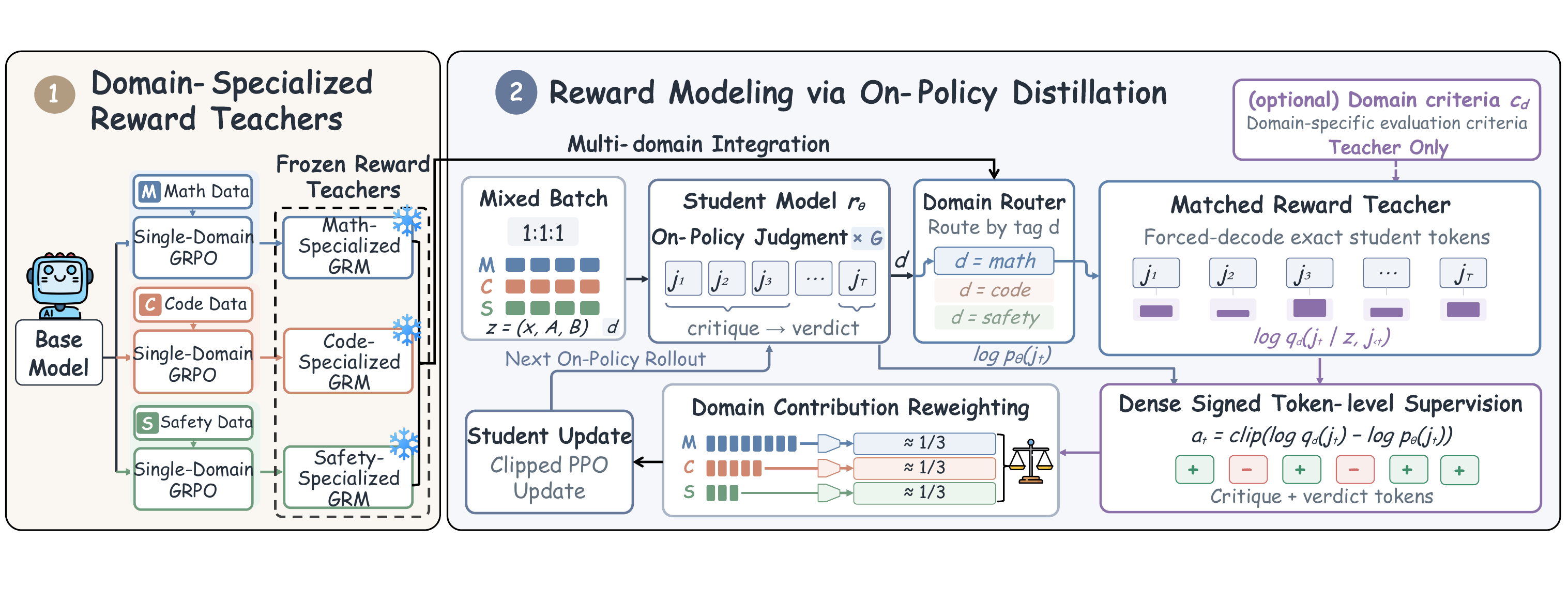}
  \vspace{-20pt}
  \caption{\textbf{Overview of \method.} Independent single-domain GRPO produces frozen math, code, and safety reward teachers. Each student-generated judgment receives token-level supervision from its domain teacher, conditioned on evaluation criteria. Domain contribution reweighting assigns target coefficients to domain token-mean losses within each micro-batch. The teachers jointly supervise one student, which performs multi-domain judgment at deployment.}
  \label{fig:overview}
\end{figure}

\method separates specialist acquisition from capability integration. Independently trained reward teachers supervise a shared student on its own judgment trajectories. Routing determines the expert distribution for each input, domain contribution reweighting sets the coefficients of domain-wise token losses, and criteria context distillation conditions the expert on explicit evaluation principles. Together, these components define the supervision used to consolidate domain expertise into one GRM (Figure~\ref{fig:overview}).

 \vspace{-20pt}

\subsection{Routed on-policy distillation}
\label{sec:method-routing}
\label{sec:method-distillation}

For each domain $d\in\mathcal D$, we train an expert $r_{T,d}$ with single-domain GRPO~\citep{shao2024deepseekmath}, using verdict correctness and a format penalty. The experts and student share a base model, tokenizer, and judgment format. We freeze experts and initialize the student from the base model to integrate multi-domain capabilities through routing-based Multi-teacher OPD (MOPD)~\citep{ma2026mopd}.

At each iteration, a fixed student snapshot $r_{\theta_{\mathrm{old}}}$ samples $G$ judgments per pairwise input. Indexing the sampled judgments by $i$, we write $j_i\sim r_{\theta_{\mathrm{old}}}(\cdot\mid z_i)$ and let $d_i$ denote the input domain. Hard routing selects $r_{T,d_i}$ to supervise every prefix of $j_i$. The student and teacher distributions are
\begin{equation}
  p_{\theta,i,t}(\cdot)=r_\theta(\cdot\mid z_i,j_{i,<t}),
  \qquad
  q_{i,t}(\cdot)=r_{T,d_i}(\cdot\mid C_{d_i}(z_i),j_{i,<t}),
  \label{eq:teacher-mixture}
\end{equation}
where $C_d(z)$ augments the judge input with optional criteria context $c_d$ (Section~\ref{sec:method-context}). This defines a domain-specific target on the reasoning states visited by the student.

We use token-mean reverse KL as the base distillation objective. For a set of sampled judgments $\mathcal B$, let $T_i=|j_i|$ and $N=\sum_{i\in\mathcal B}T_i$. With the sampled prefixes held fixed, the token-level guidance in Eq.~\ref{eq:opd} takes the form
\begin{equation}
  \mathcal L_{\mathrm{OPD}}(\theta;\mathcal B)
  =\frac{1}{N}\sum_{i\in\mathcal B}\sum_{t=1}^{T_i}
    D_{\mathrm{KL}}\!\left(p_{\theta,i,t}\Vert q_{i,t}\right).
  \label{eq:rewardopd}
\end{equation}
The local reverse-KL gradient gives a teacher--student log-ratio as the sampled-token learning signal~\citep{gu2024minillm,ma2026mopd}. With clipping threshold $A_{\max}>0$, we define
\begin{equation}
  a_{i,t}=\operatorname{clip}\!\left(
    \log q_{i,t}(j_{i,t})-\log p_{\theta,i,t}(j_{i,t}),
    -A_{\max},A_{\max}\right).
  \label{eq:advantage}
\end{equation}
Positive values favor tokens assigned higher probability by the teacher, while negative values penalize tokens overestimated by the student. The signal uses the current student distribution $p_\theta$ and a fixed teacher target $q$, providing dense supervision throughout both critique and verdict.

The domain weighting below converts these token signals into the advantages used by the student objective. We optimize a clipped PPO surrogate with importance ratios relative to $r_{\theta_{\mathrm{old}}}$. Appendix~\ref{app:theoretical-analysis} gives the local gradient derivation and the complete optimization formula.

\subsection{Domain contribution reweighting}
\label{sec:method-balance}

Judgment lengths vary across domains, so equal prompt quotas can produce imbalanced token shares in Eq.~\ref{eq:rewardopd}. Domain contribution reweighting (DCR) replaces these length-dependent shares with target coefficients for domain mean losses. Let $\alpha_d>0$ with $\sum_d\alpha_d=1$ specify the target allocation. For DCR, $\mathcal B$ is the actor micro-batch over which weights are computed. Writing $n_d=\sum_{i\in\mathcal B:d_i=d}T_i$, we define
\begin{equation}
  \hat\beta_d=\frac{n_d}{N},\qquad
  \bar\alpha_d=\frac{\alpha_d}{\sum_{e:n_e>0}\alpha_e},
  \qquad
  \kappa_i=\frac{\bar\alpha_{d_i}}{\hat\beta_{d_i}},
  \qquad
  \widehat A_{i,t}=\sg[\kappa_i a_{i,t}].
  \label{eq:domain-weights}
\end{equation}
Here $\hat\beta_d$ is the observed token share, $\bar\alpha_d$ normalizes the targets over represented domains, and $\sg$ denotes stop-gradient. Let $\ell_{i,t}(\theta)$ denote the PPO token loss before domain weighting, formed from the detached signal $a_{i,t}$, and let $\bar\ell_d(\theta)$ be its mean over the $n_d$ response tokens of domain $d$ in $\mathcal B$. The weighted loss satisfies
\begin{equation}
  \mathcal L_{\mathrm{DCR}}(\theta;\mathcal B)
  =\frac{1}{N}\sum_{i\in\mathcal B}\sum_{t=1}^{T_i}\kappa_i\ell_{i,t}(\theta)
  =\sum_{d:n_d>0}\bar\alpha_d\,\bar\ell_d(\theta).
  \label{eq:domain-token-mean}
\end{equation}

Thus, each represented domain's mean loss receives the coefficient $\bar\alpha_d$, while the token-weighted mean of $\kappa_i$ remains one. This separates the choice of domain coefficients from variations in judgment length across domains. Since the PPO token surrogate is positively homogeneous in its advantage, using $\widehat{a}_{i,t}$ implements the balanced loss weighting. More details can be found in appendix~\ref{app:theoretical-analysis}.

\subsection{Criteria context distillation}
\label{sec:method-context}

Context distillation internalizes textual guidance into model parameters~\citep{ye2026onpolicycontextdistillationlanguage,siyan2026opsd,jonas2026sdpo}. We adapt this principle to distill domain-specific evaluation criteria $c_d$ via criteria context distillation (CCD), employing an asymmetric teacher--student conditioning. Judgments are sampled as $j_i\sim r_{\theta_{\mathrm{old}}}(\cdot\mid z_i)$, while the matched teacher conditions on $C_{d_i}(z_i)$. At each fixed judgment prefix, the context-conditioned divergence is
\begin{equation}
  \mathcal K_{i,t}^{\mathrm{CCD}}(\theta)
  =D_{\mathrm{KL}}\!\left(p_{\theta,i,t}\Vert q_{i,t}\right)
  =\mathbb E_{v\sim p_{\theta,i,t}}\!\left[
    \log\frac{p_{\theta,i,t}(v)}
    {r_{T,d_i}(v\mid C_{d_i}(z_i),j_{i,<t})}
  \right],
  \label{eq:ccd-criterion}
\end{equation}
where $v$ ranges over the shared token vocabulary. Evaluating the negative log-ratio at $v=j_{i,t}$ yields the base learning signal, which is subsequently clipped (Eq.~\ref{eq:advantage}) and reweighted (Eq.~\ref{eq:domain-weights}). Thus, the student learns from criterion-conditioned teacher probabilities without explicit prompting.

The provided contexts specify derivation and constraint checks for mathematics, functional correctness for code, and harmful-assistance and over-refusal assessment for safety (Appendix~\ref{app:criteria-contexts}). When setting $c_d=\varnothing$, the mapping $C_d(z)=z$ recovers unguided distillation; otherwise, the context supplies domain-specific principles for the student to internalize. In this way, CCD internalizes explicit evaluation principles and enables a unified policy $r_\theta(\cdot\mid z)$ to apply domain-specialized criteria without any external prompts.

\section{Experiments}
\label{sec:experiments}

To evaluate how effectively \method consolidates heterogeneous evaluation capabilities into a single GRM, we conduct experiments across three distinct domains: safety, mathematics, and code. These domains require distinct evaluation criteria (intent-sensitive judgment, step-level logical verification, and execution-grounded edge-case analysis), providing a rigorous testbed for assessing cross-domain capability integration. By benchmarking \method against independently trained domain specialists, standard Mix-RL, and strong open-source GRMs, we aim to demonstrate its capacity to mitigate multi-domain performance trade-offs and improve capability coexistence.

\subsection{Experimental setup}
\label{sec:setup}

Our training corpus contains 74k cleaned pairwise preferences, organized into mathematics, code, and safety datasets. The three domain datasets are constructed primarily from the source corpora used by RM-NLHF~\citep{wang2026rmnlhf}, together with Themis-CodePreference~\citep{paul2026themis}; their full composition is provided in Appendix~\ref{app:training-data}. For both Qwen2.5-7B-Instruct~\citep{yang2025qwen25} and DeepSeek-R1-Distill-Qwen-7B~\citep{guo2025deepseekr1}, we first train one domain-specialized reward teacher on each dataset using single-domain GRPO~\citep{shao2024deepseekmath}. We then concatenate and shuffle the three datasets into a common mixed-domain training pool shared by \method and MixRL-GRPO. Training is performed on 2 nodes with 8 NVIDIA H800 GPUs each, while evaluation uses a single 8-GPU H800 node.

We evaluate \method against four categories of baselines: (1) \textbf{Base Models:} the prompted backbone models prior to reward-specific fine-tuning; (2) \textbf{Domain-Specialized RL Teachers:} models trained exclusively on a single domain via GRPO, establishing the in-domain capability ceiling; (3) \textbf{Mix-RL (MixRL-GRPO):} a generalist GRM optimized via joint GRPO on the combined multi-domain dataset, representing the prevailing capability integration paradigm; and (4) \textbf{Existing Open-Source GRMs:} a broad set of strong public reward models, including JudgeLRM~\citep{chen2025judgelrm}, CompassJudger-2~\citep{zhang2025compassjudger2}, RISE-Judge~\citep{yu2025risejudge}, RM-NLHF~\citep{wang2026rmnlhf}, RM-R1~\citep{chen2025rmr1}, and RRM~\citep{guo2025rrm}. Our evaluation spans eight public benchmark suites and 16 domain-specific subsets. Appendix~\ref{app:experimental-details} provides the full data composition, training configuration, baseline specifications, and evaluation protocol.

\subsection{Main results}
\label{sec:results}

Table~\ref{tab:main-results} reports the primary comparison on RM-Bench, RewardBench, RewardBench 2, and JudgeBench. Additional results across all 16 benchmark subsets are provided in Appendix Table~\ref{tab:additional-results}. These results reveal two notable observations of \method.

\begin{table}[t]
\centering
\caption{\textbf{Evaluation results on RM-Bench, RewardBench, RewardBench 2, and JudgeBench (\%).} Arrows ($\uparrow$) indicate that higher is better. Overall averages the displayed domains. Among model rows, \textbf{bold} denotes the best result and the second best is \underline{underlined}. Additional results and the aggregate score across all benchmark subsets are reported in Appendix Table~\ref{tab:additional-results}.}
\vspace{5pt}
\label{tab:main-results}
\scriptsize
\setlength{\tabcolsep}{1.6pt}
\renewcommand{\arraystretch}{1.09}
\begin{tabular*}{\textwidth}{@{\extracolsep{\fill}}lrrrrrrrrrrrrrr@{}}
\toprule
Model & \multicolumn{4}{c}{\textbf{RM-Bench} $\uparrow$} & \multicolumn{4}{c}{\textbf{RewardBench} $\uparrow$} & \multicolumn{3}{c}{\textbf{RewardBench 2} $\uparrow$} & \multicolumn{3}{c}{\textbf{JudgeBench} $\uparrow$} \\
\cmidrule(lr){2-5}\cmidrule(lr){6-9}\cmidrule(lr){10-12}\cmidrule(lr){13-15}
 & Safety & Math & Code & \textbf{Overall} & Safety & Math & Code & \textbf{Overall} & Safety & Math & \textbf{Overall} & Math & Code & \textbf{Overall} \\
\midrule
\rowcolor{opdblue!10}
\multicolumn{15}{c}{\textit{Base Models (LLM-as-a-Judge)}} \\
Qwen2.5-7B-Instruct & 85.13 & 59.95 & 50.24 & 65.11 & 82.57 & 62.64 & 86.38 & 77.20 & 69.78 & 39.89 & 54.84 & 55.56 & 60.27 & 57.92 \\
DeepSeek-R1-Distill-Qwen-7B & 78.28 & 82.50 & 56.73 & 72.50 & 71.89 & 86.58 & 86.79 & 81.75 & 32.67 & 57.92 & 45.30 & 67.78 & 49.32 & 58.55 \\
\addlinespace[1pt]
\rowcolor{opdblue!10}
\multicolumn{15}{c}{\textit{Existing Generative Reward Models}} \\
JudgeLRM-7B & 87.23 & 60.45 & 51.75 & 66.48 & 80.41 & 64.43 & 80.08 & 74.97 & 68.67 & 44.26 & 56.47 & 67.78 & 60.27 & 64.03 \\
CompassJudger-2-7B-Instruct & 90.02 & 62.38 & 54.68 & 69.03 & 87.30 & 85.68 & 91.26 & 88.08 & 78.44 & 44.26 & 61.35 & 64.44 & 60.27 & 62.36 \\
RISE-Judge-Qwen2.5-7B & 90.70 & 59.84 & 55.85 & 68.80 & 88.24 & \textbf{99.33} & 89.74 & \textbf{92.44} & 82.44 & 54.64 & 68.54 & 70.00 & 57.53 & 63.77 \\
RM-NLHF-Qwen-7B & 82.39 & 77.38 & 58.19 & 72.65 & 76.76 & 91.95 & 93.80 & 87.50 & 72.67 & 67.76 & 70.22 & 72.22 & 64.38 & 68.30 \\
RM-R1-Qwen2.5-Instruct-7B & 91.21 & 63.26 & 52.53 & 69.00 & 88.11 & 81.21 & 89.43 & 86.25 & 72.67 & 55.74 & 64.21 & 68.89 & 54.79 & 61.84 \\
RM-R1-DS-Distilled-Qwen-7B & 84.28 & 84.12 & 56.48 & 74.96 & 79.32 & 90.60 & 86.99 & 85.64 & 52.22 & 67.76 & 59.99 & 71.11 & 58.90 & 65.01 \\
RRM-7B & 83.45 & 77.00 & 57.02 & 72.49 & 77.43 & 83.67 & 89.84 & 83.65 & 52.67 & 68.85 & 60.76 & 68.89 & 63.01 & 65.95 \\
\addlinespace[1pt]
\rowcolor{opdblue!10}
\multicolumn{15}{c}{\textit{Our Methods (Qwen2.5-7B-Instruct)}} \\
RL Teacher & \underline{96.12} & 65.18 & 56.43 & 72.58 & \underline{92.70} & 89.04 & 94.21 & \underline{91.98} & \textbf{84.67} & 62.84 & \underline{73.76} & \underline{77.78} & 61.64 & 69.71 \\
\rowcolor{opdorange!8}
MixRL-GRPO & 93.93 & 62.70 & 53.70 & 70.11 & 92.16 & 83.67 & 91.57 & 89.13 & 76.22 & 46.99 & 61.61 & 62.22 & 53.42 & 57.82 \\
\rowcolor{opdgreen!9}
\textbf{RM-OPD} & \textbf{96.75} & 64.76 & 53.41 & 71.64 & \textbf{93.92} & 87.92 & 92.58 & 91.47 & \underline{84.00} & 54.10 & 69.05 & 75.56 & 63.01 & 69.29 \\
\rowcolor{opdgreen!9}
$\Delta$ (vs.\ MixRL) & \textcolor{opdgreen}{+2.82} & \textcolor{opdgreen}{+2.06} & \textcolor{draftred}{-0.29} & \textcolor{opdgreen}{+1.53} & \textcolor{opdgreen}{+1.76} & \textcolor{opdgreen}{+4.25} & \textcolor{opdgreen}{+1.01} & \textcolor{opdgreen}{+2.34} & \textcolor{opdgreen}{+7.78} & \textcolor{opdgreen}{+7.11} & \textcolor{opdgreen}{+7.44} & \textcolor{opdgreen}{+13.34} & \textcolor{opdgreen}{+9.59} & \textcolor{opdgreen}{+11.47} \\
\addlinespace[1pt]
\rowcolor{opdblue!10}
\multicolumn{15}{c}{\textit{Our Methods (DeepSeek-R1-Distill-Qwen-7B)}} \\
RL Teacher & 91.81 & \underline{90.32} & \textbf{61.89} & \textbf{81.34} & 83.92 & \underline{96.42} & \underline{94.82} & 91.72 & 62.44 & \underline{78.69} & 70.57 & \textbf{82.22} & \textbf{68.49} & \textbf{75.36} \\
\rowcolor{opdorange!8}
MixRL-GRPO & 89.64 & 87.86 & 58.58 & 78.69 & 82.16 & 92.62 & 93.80 & 89.53 & 62.89 & 74.32 & 68.61 & 72.22 & \underline{65.75} & 68.99 \\
\rowcolor{opdgreen!9}
\textbf{RM-OPD} & 92.01 & \textbf{91.30} & \underline{60.62} & \underline{81.31} & 84.05 & 95.75 & \textbf{95.02} & 91.61 & 70.22 & \textbf{79.23} & \textbf{74.73} & \underline{77.78} & \textbf{68.49} & \underline{73.14} \\
\rowcolor{opdgreen!9}
$\Delta$ (vs.\ MixRL) & \textcolor{opdgreen}{+2.37} & \textcolor{opdgreen}{+3.44} & \textcolor{opdgreen}{+2.04} & \textcolor{opdgreen}{+2.62} & \textcolor{opdgreen}{+1.89} & \textcolor{opdgreen}{+3.13} & \textcolor{opdgreen}{+1.22} & \textcolor{opdgreen}{+2.08} & \textcolor{opdgreen}{+7.33} & \textcolor{opdgreen}{+4.91} & \textcolor{opdgreen}{+6.12} & \textcolor{opdgreen}{+5.56} & \textcolor{opdgreen}{+2.74} & \textcolor{opdgreen}{+4.15} \\
\bottomrule
\end{tabular*}
\vspace{-10pt}
\end{table}

\FloatBarrier

\textbf{\method mitigates the performance limitations of Mix-RL.} 
A common bottleneck in multi-domain GRMs is that Mix-RL typically underperforms independently trained specialists due to gradient conflicts (the seesaw effect). Our results demonstrate that \method effectively addresses this issue. Across both Qwen2.5 and DeepSeek-R1-Distill bases, MixRL-GRPO consistently lags behind the single-domain RL Teachers (e.g., showing noticeable gaps on RewardBench 2 and JudgeBench). In contrast, \method reliably matches the teachers or narrows the gap across domains. Notably, \method exhibits a \emph{teacher-surpassing} effect on complex reasoning tasks: it outperforms the specialized teacher on RewardBench 2 and RM-Bench Safety. This validates the hypothesis that separating capability acquisition from integration via on-policy distillation mitigates mutual degradation and facilitates positive cross-domain synergy.

\textbf{\method consistently outperforms existing mixed-domain GRM baselines.} 
Beyond matching specialist ceilings, \method demonstrates empirical advantages over other generalist GRMs built by mixed-domain training. When controlling for the base model family, DeepSeek \method improves the overall macro average across all four core benchmarks compared to the matched RM-R1 and RM-NLHF variants. Furthermore, as detailed in Appendix Table~\ref{tab:additional-results}, Qwen \method achieves an average accuracy of 75.41 across all 16 subsets, outperforming existing open-source GRMs, including RISE-Judge (74.12), RM-NLHF (71.76), RM-R1 (71.22), and RRM (68.88). The fact that \method achieves these results without scaling up model size or data volume indicates that the improvements are primarily attributed to the architectural design of routed expert distillation over standard joint multi-domain optimization.

\section{Analysis}
\label{sec:analysis}

To clarify the efficacy of \method and unveil the limitations of Mix-RL, we now conduct a granular analysis to isolate its underlying mechanisms. Specifically, we investigate how it breaks the multi-domain coexistence bottleneck, quantify the contributions of individual components, and evaluate its out-of-domain generalization. Unless noted otherwise, all analysis utilizes DeepSeek-R1-Distill-Qwen-7B, supplemented by qualitative case studies in Appendix~\ref{app:case-studies}.

\subsection{Multi-domain scaling and coexistence}
\label{sec:domain-scaling}
\label{sec:domain-ablation}

\begin{figure}[!htbp]
  \centering
  \includegraphics[width=\textwidth]{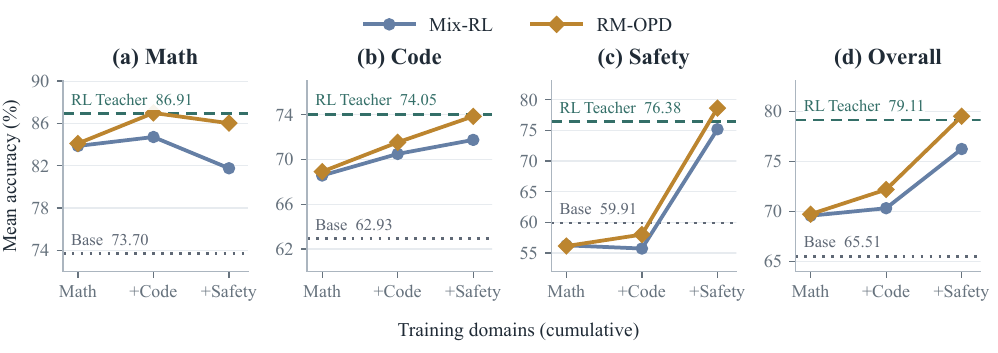}
  \vspace{-25pt}
  \caption{\textbf{Multi-domain scaling.} Domain panels average four fixed subsets each; Overall averages all 12 across every configuration. Training data expands from Math to Math+Code to Math+Code+Safety. Horizontal lines mark the performance of Base and domain specialists.}
  \label{fig:domain-scaling}
\end{figure}

A fundamental challenge in training generalist GRMs is the \emph{seesaw effect} induced by cross-domain interference, where jointly optimizing heterogeneous domains degrades existing capabilities. Figure~\ref{fig:domain-scaling} evaluates capability coexistence by scaling domains from \emph{Math} to \emph{+Code} and \emph{+Safety}, comparing \method against the mixed-domain RL baseline (Mix-RL). Each domain averages four fixed evaluation subsets (Appendix~\ref{app:domain-scaling}).We highlight two critical observations from the scaling experiment.

\vspace{-10pt}
\FloatBarrier
\vspace{-5pt}

\textbf{Mix-RL suffers from seesaw effect and exhibits consistent performance bottlenecks as domain scales.} While adding the code domain is relatively benign, integrating safety introduces conflicting evaluation objectives. Consequently, Mix-RL's mathematical accuracy drops by 2.96 points, falling below its single-domain baseline and our RM-OPD(Figure~\ref{fig:domain-scaling}a). This degradation exemplifies the cross-domain interference inherent in joint sparse-reward optimization, which consistently limits the model's overall performance ceiling as domain scales.

\textbf{\method mitigates cross-domain interference and coexists heterogeneous capabilities better.} \method preserves its math capability (only a 0.97 drop) when safety is introduced. Most notably, as domain scales, \method does not merely avoid forgetting; the full model (\emph{Math+Code+Safety}) actually \emph{outperforms the single-domain RL Teacher specialists} in Safety (Figure~\ref{fig:domain-scaling}c) and Overall average (Figure~\ref{fig:domain-scaling}d). By shifting from joint sparse-reward RL to dense on-policy distillation, \method effectively decouples capability acquisition from inter-domain conflicts, allowing heterogeneous judgement skills to coexist in a higher level.

\subsection{Ablation study}
\label{sec:analysis-ablation}
\label{sec:analysis-ccd}

\vspace{-5pt}
\noindent
\begin{minipage}[t]{0.50\textwidth}
\vspace{0pt}
\paragraph{Domain contribution reweighting.}

As we observed, domain contributions are imbalanced: mathematics, code, and safety account for $63.28\%$, $21.96\%$, and $14.76\%$ of generated tokens, respectively. Unequal supervision shares give math the largest coefficient in token-averaged gradient aggregation. Removing DCR reduces code and safety accuracy by 3.43 and 2.48 points while leaving mathematics nearly unchanged (Table~\ref{tab:dcr-ablation-summary}; Appendix~\ref{app:dcr-ablation}). Reweighting these contributions thus supports better coexistence of domain capabilities.

\end{minipage}
\hfill
\begin{minipage}[t]{0.47\textwidth}
\vspace{-10pt}
\centering
\makeatletter\def\@captype{table}\makeatother
\caption{\textbf{DCR ablation (\%).} Domain means over 12 benchmark subsets.}
\vspace{4pt}
\label{tab:dcr-ablation-summary}
\small
\setlength{\tabcolsep}{3.0pt}
\renewcommand{\arraystretch}{1.06}
\begin{tabular}{@{}lrrrr@{}}
  \toprule
  Model & Math & Code & Safety & Avg. \\
  \midrule
  Base & 73.70 & 62.93 & 59.91 & 65.51 \\
  RL Teacher & 86.91 & 74.05 & 76.38 & 79.11 \\
  \midrule
  \rowcolor{opdgreen!9}
  \multicolumn{5}{l}{\textbf{RM-OPD}} \\
  w/ DCR & 86.02 & \textbf{73.85} & \textbf{78.60} & \textbf{79.49} \\
  w/o DCR & \textbf{86.11} & 70.42 & 76.13 & 77.55 \\
  $\Delta$ (vs.\ w/ DCR) & \textcolor{opdgreen}{+0.09} & \textcolor{draftred}{-3.43} & \textcolor{draftred}{-2.48} & \textcolor{draftred}{-1.94} \\
  \bottomrule
\end{tabular}
\end{minipage}

\begin{figure}[!htbp]
  \centering
  \includegraphics[width=\textwidth,, trim=3pt 10pt 3pt 5pt, clip]{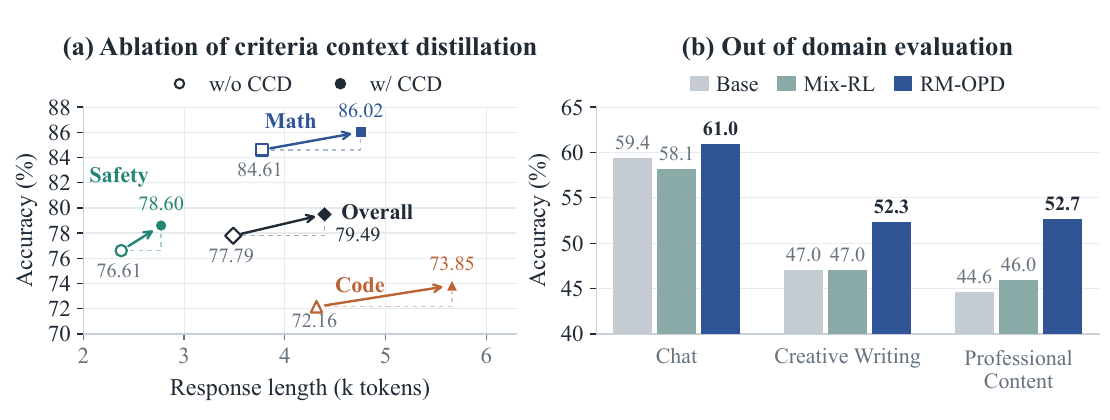}
  \vspace{-16pt}
  \caption{\textbf{Ablation of criteria context distillation and out of domain evaluation.} \textbf{(a)} Accuracy and response length without CCD and with CCD. Each domain averages four subsets, and Overall averages all 12 benchmark subsets. \textbf{(b)} For OOD generalization, we evaluate Base, Mix-RL, and RM-OPD on RM-Bench Chat/Normal and two RMB-BoN generation subsets.}
  \label{fig:ccd-displacement}
\end{figure}

\vspace{-5pt}
\paragraph{Criteria context distillation.}
Figure~\ref{fig:ccd-displacement}(a) shows that CCD improves all three domain scores. The gain is largest in safety, followed by code and math, indicating that explicit teacher-side criteria sharpen domain-specific supervision rather than benefiting a single benchmark.The accuracy gains are accompanied by longer judgments across all 12 subsets. Average response length increases from 3,488 to 4,395 tokens, while overall accuracy improves by  1.70 points. Thus, teacher-side criteria context is associated with both more extensive verification and more accurate final verdicts.

\vspace{-10pt}

\vspace{-10pt}
\subsection{Out of domain Generalization}
\label{sec:analysis-transfer}

Figure~\ref{fig:ccd-displacement}(b) evaluates generalization to chat and content-generation tasks beyond the three specialist training domains. RM-OPD achieves 60.98\% accuracy on RM-Bench Chat/Normal, compared with 59.43\% for Base and 58.14\% for Mix-RL. On RMB-BoN, creative-writing and professional-content evaluation reach 52.32\% and 52.70\%, exceeding Mix-RL by 5.30 and 6.75 points, respectively. These results show positive transfer on the displayed tasks. The gains depend on the evaluation setting: on RMB pairwise creative writing, RM-OPD scores 64.96\% versus 67.72\% for Mix-RL. 

\FloatBarrier

\section{Conclusion}

We identified the structural limitations of Mixed-domain Reinforcement Learning (Mix-RL) in training general-purpose Generative Reward Models (GRMs), where sparse rewards and cross-domain interference constrain the attainable performance ceiling. To address these bottlenecks, we introduced \method, a two-stage framework that decouples domain specialist acquisition from capability integration for GRMs. By employing routed on-policy distillation, domain contribution reweighting, and criteria context distillation, \method provides dense, token-level supervision to effectively consolidate heterogeneous evaluation capabilities into a generalist GRM. Extensive experiments demonstrate that \method consistently outperforms Mix-RL baselines and exhibits an emergent teacher-surpassing effect across diverse domains. Ultimately, our work establishes a highly effective and promising paradigm for developing robust generalist reward models.

\bibliography{iclr2027_conference}
\bibliographystyle{iclr2027_conference}

\appendix
\clearpage

\section{Theoretical Analysis and Implementation Details}
\label{app:theoretical-analysis}

\paragraph{Prompt quotas and teacher scoring.}
Integer prompt quotas use largest-remainder allocation, with shuffled domain pools recycled when exhausted. For each prompt, the rollout policy samples $G$ judgments; each is scored only by its domain teacher. Teacher services return causal log-probabilities of the sampled tokens. In padded tensors, the response mask $m_{i,t}\in\{0,1\}$ selects generated response positions, excluding prompt and padding tokens. Thus, $T_i=\sum_t m_{i,t}$, and the response-token sums in Section~\ref{sec:method} are implemented as masked sums below. New rollouts use the student parameters after each synchronous actor update.

\paragraph{Criteria context and token alignment.}
The input transformation $C_d$ inserts the domain criteria after the leading chat-template tokens during teacher scoring. We remove the inserted positions from the returned score tensors to align teacher log-probabilities with the student's sampled tokens. Criteria context can be enabled independently by domain; the default configuration enables it for all three domains. Student rollout and evaluation inputs use the original pairwise prompt.

\paragraph{Local reverse-KL gradient.}
At a fixed student-visited prefix, abbreviate $p_{\theta,i,t}$ and $q_{i,t}$ as $p_\theta$ and $q$. With frozen $q$ and $p_{\mathrm{old}}=p_{\theta_{\mathrm{old}}}$, the local reverse-KL gradient satisfies
\begin{equation}
  \begin{aligned}
  \nabla_\theta D_{\mathrm{KL}}(p_\theta\Vert q)
  &=\mathbb E_{v\sim p_\theta}
    \left[\log\frac{p_\theta(v)}{q(v)}\nabla_\theta\log p_\theta(v)\right]\\
  &=\mathbb E_{v\sim p_{\mathrm{old}}}
    \left[\frac{p_\theta(v)}{p_{\mathrm{old}}(v)}
    \log\frac{p_\theta(v)}{q(v)}\nabla_\theta\log p_\theta(v)\right].
  \end{aligned}
  \label{eq:local-kl-gradient}
\end{equation}
The first equality uses $\sum_v\nabla_\theta p_\theta(v)=0$; the second changes the sampling distribution to the rollout policy. For $A^0(v)=\sg[\log q(v)-\log p_\theta(v)]$, differentiating $-[p_\theta(v)/p_{\mathrm{old}}(v)]A^0(v)$ gives the corresponding sampled-token gradient. This supplies the policy-gradient form used in \method. The actor recomputes current student log-probabilities, clips the signed log-ratio as in Eq.~\ref{eq:advantage}, applies domain weights, and detaches the advantages before the PPO update. The sampled prefixes and teacher scores remain fixed throughout this computation.

\paragraph{Domain contribution reweighting.}
For micro-batch $\mathcal B_b$, write $N_b$ for its response-token count, $n_{d,b}$ for its domain counts, and $\bar\alpha_{d,b}$ for its normalized targets. Let $\bar\ell_{d,b}$ be the mean unweighted token loss of domain $d$ in that micro-batch. With $N_{\mathrm{all}}=\sum_b N_b$, global token-mean accumulation gives
\begin{equation}
  \mathcal L_{\mathrm{weighted}}
  =\sum_b\frac{N_b}{N_{\mathrm{all}}}
    \sum_{d:n_{d,b}>0}\bar\alpha_{d,b}\bar\ell_{d,b}.
  \label{eq:domain-global-aggregation}
\end{equation}
Weights are computed after dynamic micro-batching and reused for all tokens of the same judgment. The DeepSeek-based configuration uses an 8,192-token budget per GPU with sequence parallelism of two for dynamic micro-batching. For $\kappa_i>0$, both the PPO minimum and the negative-advantage dual-clip maximum commute with multiplication by $\kappa_i$. Weighting the detached advantage therefore implements Eq.~\ref{eq:domain-token-mean} for the token surrogate below. Single-domain micro-batches have unit weights.

\paragraph{Exact clipped surrogate.}
The rollout-to-actor importance ratio is $\rho_{i,t}=p_{\theta,i,t}(j_{i,t})/p_{\theta_{\mathrm{old}},i,t}(j_{i,t})$. The implementation bounds its logarithm for numerical stability and applies PPO clipping:
\begin{equation}
  \begin{aligned}
  \widetilde\rho_{i,t}
  &=\exp\!\left(\operatorname{clip}(\log\rho_{i,t},-20,20)\right),\\
  u_{i,t}&=\min\!\left(
    \widetilde\rho_{i,t}\widehat A_{i,t},
    \operatorname{clip}(\widetilde\rho_{i,t},1-\epsilon,1+\epsilon)\widehat A_{i,t}
  \right).
  \end{aligned}
  \label{eq:clipped-surrogate}
\end{equation}
The implemented loss additionally applies dual clipping to negative advantages:
\begin{equation}
  \mathcal L_{\mathrm{RM\text{-}OPD}}
  =-\frac{1}{N_{\mathrm{all}}}\sum_b\sum_{i\in\mathcal B_b,t}m_{i,t}
  \begin{cases}
    \max(u_{i,t},c\widehat A_{i,t}), & \widehat A_{i,t}<0,\\
    u_{i,t}, & \widehat A_{i,t}\geq0,
  \end{cases}
  \label{eq:pgloss}
\end{equation}
Here $\widehat A_{i,t}$ uses the weights computed in the micro-batch containing judgment $i$, and $N_{\mathrm{all}}$ counts response tokens across the complete optimizer update. The unweighted token loss $\ell_{i,t}(\theta)$ in Eq.~\ref{eq:domain-token-mean} is the negative of the corresponding branch in Eq.~\ref{eq:pgloss}, with $\widehat A_{i,t}$ replaced by $\sg[a_{i,t}]$. We use PPO clipping radius $\epsilon=0.2$, dual-clip factor $c=3$, and advantage threshold $A_{\max}=10$. These clipping operations stabilize the local policy update. The default student objective consists of this distillation surrogate; outcome and format rewards are logged separately, and the token advantages are used directly without group reward normalization.

\section{Experimental Details}
\label{app:experimental-details}

\subsection{Training configuration}
\label{app:training-configuration}

Table~\ref{tab:training-hyperparameters} summarizes the key hyperparameters for student distillation from DeepSeek-R1-Distill-Qwen-7B and Qwen2.5-7B-Instruct. We use synchronous training in VeRL with an FSDP actor and vLLM for generation and teacher scoring. Training runs on two eight-GPU H800 nodes.

\begin{table}[htbp]
  \centering
  \caption{\textbf{Student training hyperparameters.} Configuration for \method with DeepSeek-R1-Distill-Qwen-7B and Qwen2.5-7B-Instruct.}
  \label{tab:training-hyperparameters}
  \vspace{5pt}
  \normalsize
  \setlength{\tabcolsep}{8pt}
  \renewcommand{\arraystretch}{1.12}
  \begin{tabular*}{0.82\textwidth}{@{\extracolsep{\fill}}lc@{}}
    \toprule
    \textbf{Hyperparameter} & \textbf{Value} \\
    \midrule
    \multicolumn{2}{l}{\textit{Data and sampling}} \\
    \midrule
    Maximum prompt length & 6,144 tokens \\
    Maximum response length & 8,192 tokens \\
    Training batch size & 128 prompts \\
    Rollouts per prompt & 8 \\
    Sampling temperature & 1.0 \\
    Sampling top-$p$ & 1.0 \\
    \midrule
    \multicolumn{2}{l}{\textit{Optimization}} \\
    \midrule
    Optimizer & AdamW \\
    Learning rate & $1\times10^{-6}$ \\
    PPO epochs per batch & 1 \\
    PPO clipping threshold ($\epsilon$) & 0.2 \\
    Dual-clip factor ($c$) & 3.0 \\
    Maximum training steps & 1,000 \\
    \midrule
    \multicolumn{2}{l}{\textit{Distillation}} \\
    \midrule
    Distillation objective & Reverse KL \\
    Token-advantage clipping ($A_{\max}$) & 10.0 \\
    Target domain ratio (math/code/safety) & $1:1:1$ \\
    Domain contribution reweighting & Enabled \\
    Criteria context distillation & Enabled \\
    \bottomrule
  \end{tabular*}
\end{table}

The student is optimized with the token-mean policy-gradient surrogate in Eq.~\ref{eq:pgloss}; outcome and format rewards are logged for monitoring, with no additional task-reward, entropy, or reference-model KL terms in the student objective. Prompt quotas enforce the target domain ratio, and DCR reweights domain losses within each micro-batch. CCD conditions each teacher on its domain criteria using a 256-token prefix budget, while the student receives the original judgment prompt. Teachers return log-probabilities for the student's sampled tokens without materializing full-vocabulary logits.
\FloatBarrier

\subsection{Training dataset}
\label{app:training-data}

We exclusively use pairwise preference data. The student training pool concatenates the mathematics, code, and safety datasets used to train the three specialist reward teachers and then globally shuffles them with seed 42. Table~\ref{tab:training-data} gives the resulting source distribution. The three domains are nearly balanced in the pool, while the domain-weighted sampler enforces prompt-level batch quotas of 1:1:1. Preference labels supervise the specialists; during distillation, the student instead learns from routed teacher token probabilities on its own sampled judgments. No offline teacher traces or additional critique SFT data are used.

\begin{table}[H]
  \centering
  \caption{\textbf{Training data overview.} Counts are preference pairs in the student prompt pool. Source names denote the provenance stored with each example; ``Skywork mined'' comprises domain-filtered subsets of the Skywork preference mixture~\citep{liu2025skyworkv2}.}
  \vspace{5pt}
  \label{tab:training-data}
  \small
  \begin{tabular}{lr}
    \toprule
    Source & Size \\
    \midrule
    \multicolumn{2}{c}{\textit{Mathematics (25,000)}} \\
    \midrule
    math\_rlvr mixture & 24,277 \\
    Skywork Magpie Math & 619 \\
    HelpSteer3 STEM (math-filtered) & 104 \\
    \midrule
    \multicolumn{2}{c}{\textit{Code (24,201)}} \\
    \midrule
    Themis-CodePreference & 9,501 \\
    Skywork general code mined & 5,525 \\
    HelpSteer3 Code & 5,004 \\
    Code-Preference-Pairs & 4,171 \\
    \midrule
    \multicolumn{2}{c}{\textit{Safety (24,885)}} \\
    \midrule
    WildGuardMix & 10,214 \\
    hh-rlhf-safety-v3-dpo & 8,756 \\
    Skywork WildGuard & 5,915 \\
    \midrule
    \textbf{Total} & \textbf{74,086} \\
    \bottomrule
  \end{tabular}
\end{table}

The mathematics mixture contains 24,277 verifier-filtered RLVR pairs drawn from AceReason-Math, DAPO-Math, omega-combined, MathSub-30K, and rlvr\_orz\_math, plus smaller Skywork and HelpSteer3 subsets. Code combines four complementary sources to increase task and programming-language diversity. Safety combines harmful-request preference pairs with anti-over-refusal examples. Across the pool, A/B order is randomized approximately evenly (36,767 swapped and 37,319 unswapped examples), and no domain identifier is exposed in the judge prompt.

\subsection{Evaluation benchmarks and metrics}
\label{app:evaluation-benchmarks}

We evaluate on eight public reward-model or LLM-judge suites: RewardBench~\citep{lambert2025rewardbench}, RewardBench 2~\citep{malik2026rewardbench2}, RM-Bench~\citep{liu2025rmbench}, RMB and RMB-BoN~\citep{zhou2025rmb}, JudgeBench~\citep{tan2024judgebench}, PKU-SafeRLHF~\citep{ji2024pkusaferlhf}, and Themis-CodeRewardBench~\citep{paul2026themis}. Table~\ref{tab:evaluation-data} reports counts under the official sample-group definition. The final evaluation contains 35,184 official samples across 16 benchmark subsets spanning mathematics, code, and safety.

\begin{table}[H]
  \centering
  \caption{\textbf{Evaluation benchmark overview.} Domain columns count the official sample groups used in the final evaluation.}
  \vspace{5pt}
  \label{tab:evaluation-data}
  \small
  \begin{tabular}{lrrrr}
    \toprule
    Benchmark & Math & Code & Safety & Total \\
    \midrule
    RewardBench & 447 & 984 & 740 & 2,171 \\
    RewardBench 2 & 183 & -- & 450 & 633 \\
    RM-Bench & 4,761 & 2,052 & 3,969 & 10,782 \\
    RMB & -- & 1,224 & 7,064 & 8,288 \\
    RMB-BoN & -- & 280 & 1,677 & 1,957 \\
    JudgeBench & 90 & 73 & -- & 163 \\
    PKU-SafeRLHF & -- & -- & 2,324 & 2,324 \\
    Themis-CodeRewardBench & -- & 8,866 & -- & 8,866 \\
    \midrule
    \textbf{Total} & \textbf{5,481} & \textbf{13,479} & \textbf{16,224} & \textbf{35,184} \\
    \bottomrule
  \end{tabular}
\end{table}

All datasets are converted to one pairwise judge format with randomized A/B order and hidden domain labels. RewardBench 2 Best-of-4 examples are expanded into three comparisons, and RMB-BoN compares the preferred response against every remaining candidate. An official sample is correct only if every expanded comparison is correct. RM-Bench instead retains its official $3\times3$ style grid and computes hard, normal, and easy accuracy from the corresponding cells.

Each displayed benchmark subset reports the accuracy scores. For a benchmark with multiple displayed domains, Overall is their unweighted arithmetic mean. The final Average is the macro average over the 16 benchmark subsets and does not count Overall columns. In Tables~\ref{tab:main-results} and~\ref{tab:additional-results}, summaries are computed from the displayed subset accuracies and rounded to two decimal places; differences subtract the displayed model scores. Sampling uses temperature 0.6, top-$p=1$, a 6,144-token prompt limit and an 8,192-token response limit. The same judgment parser is used for training-time monitoring and final evaluation.

\paragraph{Models and baselines.}
The two student initializations are Qwen2.5-7B-Instruct~\citep{yang2025qwen25} and DeepSeek-R1-Distill-Qwen-7B~\citep{guo2025deepseekr1}. The comparison set contains JudgeLRM-7B~\citep{chen2025judgelrm}, CompassJudger-2-7B-Instruct~\citep{zhang2025compassjudger2}, RISE-Judge-Qwen2.5-7B~\citep{yu2025risejudge}, RM-NLHF-Qwen-7B~\citep{wang2026rmnlhf}, RM-R1-DeepSeek-Distilled-Qwen-7B and RM-R1-Qwen2.5-Instruct-7B~\citep{chen2025rmr1}, and RRM-7B~\citep{guo2025rrm}. Each base family additionally has its own MixRL-GRPO, RL Teacher, and \method rows. We instantiate Mix-RL as MixRL-GRPO~\citep{shao2024deepseekmath}, which trains one model on the pooled safety, mathematics, and code data using the same outcome and format rewards as the domain specialists. MixRL-GRPO and \method use the same base initialization and training pool, controlling for initialization and data while contrasting sparse outcome optimization with routed on-policy distillation. Table labels are shortened where needed. 

\subsection{Additional benchmark results}

Table~\ref{tab:additional-results} complements the main comparison with RMB, RMB-BoN, PKU-SafeRLHF, and Themis, and reports the macro average over all 16 benchmark subsets.

\begin{table}[H]
\centering
\caption{\textbf{Additional evaluation results (\%).} Overall is the unweighted mean of the displayed domains within each benchmark, while Average equally weights all 16 benchmark subsets across Table~\ref{tab:main-results} and this table. Among model rows, bold and underlined values mark the best and second-best distinct results. Delta rows report RM-OPD minus MixRL-GRPO, with positive values in \textcolor{opdgreen}{green} and negative values in \textcolor{draftred}{red}.}
\vspace{5pt}
\label{tab:additional-results}
\scriptsize
\setlength{\tabcolsep}{1.9pt}
\renewcommand{\arraystretch}{1.09}
\begin{tabular*}{\textwidth}{@{\extracolsep{\fill}}lrrrrrrrrr@{}}
\toprule
Model & \multicolumn{3}{c}{\textbf{RMB} $\uparrow$} & \multicolumn{3}{c}{\textbf{RMB-BoN} $\uparrow$} & \multicolumn{1}{c}{\textbf{PKU-SafeRLHF} $\uparrow$} & \multicolumn{1}{c}{\textbf{Themis} $\uparrow$} & \textbf{Average} $\uparrow$ \\
\cmidrule(lr){2-4}\cmidrule(lr){5-7}\cmidrule(lr){8-8}\cmidrule(lr){9-9}
 & Safety & Code & \textbf{Overall} & Safety & Code & \textbf{Overall} & Safety & Code & All 16 \\
\midrule
\rowcolor{opdblue!10}
\multicolumn{10}{c}{\textit{Base Models (LLM-as-a-Judge)}} \\
Qwen2.5-7B-Instruct & 78.38 & 77.04 & 77.71 & 61.54 & 57.86 & 59.70 & 65.96 & 67.72 & 66.31 \\
DeepSeek-R1-Distill-Qwen-7B & 63.46 & 70.42 & 66.94 & 45.32 & 43.93 & 44.63 & 56.80 & 58.89 & 63.08 \\
\addlinespace[1pt]
\rowcolor{opdblue!10}
\multicolumn{10}{c}{\textit{Existing Generative Reward Models}} \\
JudgeLRM-7B & 78.43 & 78.27 & 78.35 & 63.27 & 63.57 & 63.42 & 65.62 & 64.40 & 67.43 \\
CompassJudger-2-7B-Instruct & \underline{80.29} & 83.33 & 81.81 & \textbf{68.16} & 71.07 & \underline{69.62} & 68.03 & 70.05 & 72.48 \\
RISE-Judge-Qwen2.5-7B & \textbf{80.39} & 83.99 & \underline{82.19} & \underline{66.49} & 71.43 & 68.96 & 67.13 & 68.13 & 74.12 \\
RM-NLHF-Qwen-7B & 70.44 & 78.19 & 74.32 & 52.95 & 61.43 & 57.19 & 61.88 & 65.80 & 71.76 \\
RM-R1-Qwen2.5-Instruct-7B & 75.71 & 82.03 & 78.87 & 60.17 & 70.71 & 65.44 & 64.97 & 68.15 & 71.22 \\
RM-R1-DS-Distilled-Qwen-7B & 68.04 & 74.18 & 71.11 & 48.48 & 53.57 & 51.03 & 57.62 & 59.69 & 68.34 \\
RRM-7B & 68.96 & 77.04 & 73.00 & 52.18 & 62.14 & 57.16 & 57.44 & 62.55 & 68.88 \\
\addlinespace[1pt]
\rowcolor{opdblue!10}
\multicolumn{10}{c}{\textit{Our Methods (Qwen2.5-7B-Instruct)}} \\
RL Teacher & 78.95 & \textbf{86.11} & \textbf{82.53} & 65.18 & \textbf{76.07} & \textbf{70.63} & 70.91 & \textbf{80.72} & \textbf{77.41} \\
\rowcolor{opdorange!8}
MixRL-GRPO & 72.18 & 82.52 & 77.35 & 53.01 & 71.07 & 62.04 & \underline{71.47} & 76.31 & 71.45 \\
\rowcolor{opdgreen!9}
\textbf{RM-OPD} & 73.85 & \underline{84.72} & 79.29 & 55.99 & \underline{74.64} & 65.32 & \textbf{72.12} & \underline{79.22} & 75.41 \\
\rowcolor{opdgreen!9}
$\Delta$ (vs.\ MixRL) & \textcolor{opdgreen}{+1.67} & \textcolor{opdgreen}{+2.20} & \textcolor{opdgreen}{+1.94} & \textcolor{opdgreen}{+2.98} & \textcolor{opdgreen}{+3.57} & \textcolor{opdgreen}{+3.28} & \textcolor{opdgreen}{+0.65} & \textcolor{opdgreen}{+2.91} & \textcolor{opdgreen}{+3.96} \\
\addlinespace[1pt]
\rowcolor{opdblue!10}
\multicolumn{10}{c}{\textit{Our Methods (DeepSeek-R1-Distill-Qwen-7B)}} \\
RL Teacher & 66.19 & 78.51 & 72.35 & 46.93 & 60.71 & 53.82 & 67.34 & 70.98 & 75.11 \\
\rowcolor{opdorange!8}
MixRL-GRPO & 67.33 & 75.49 & 71.41 & 48.90 & 57.14 & 53.02 & 65.92 & 68.86 & 72.72 \\
\rowcolor{opdgreen!9}
\textbf{RM-OPD} & 72.18 & 76.96 & 74.57 & 54.32 & 59.64 & 56.98 & 68.12 & 71.27 & \underline{76.06} \\
\rowcolor{opdgreen!9}
$\Delta$ (vs.\ MixRL) & \textcolor{opdgreen}{+4.85} & \textcolor{opdgreen}{+1.47} & \textcolor{opdgreen}{+3.16} & \textcolor{opdgreen}{+5.42} & \textcolor{opdgreen}{+2.50} & \textcolor{opdgreen}{+3.96} & \textcolor{opdgreen}{+2.20} & \textcolor{opdgreen}{+2.41} & \textcolor{opdgreen}{+3.34} \\
\bottomrule
\end{tabular*}
\end{table}

\FloatBarrier

\subsection{Domain-wise results for multi-domain scaling}
\label{app:domain-scaling}

Table~\ref{tab:domain-scaling} reports the means underlying Figure~\ref{fig:domain-scaling}. Mathematics averages RM-Bench, RewardBench, RewardBench 2, and JudgeBench; code averages RM-Bench, RewardBench, JudgeBench, and Themis; safety averages RM-Bench, RewardBench, RewardBench 2, and PKU-SafeRLHF. The three configurations expand the training-domain set while retaining the same evaluation subsets. For \method, the active specialist pool follows that set. Base is the prompted initialization, and each RL Teacher column uses the specialist trained for that domain. 

\vspace{-10pt}
\begin{table}[!htbp]
  \centering
  \caption{\textbf{Domain means under multi-domain scaling (\%).} Each domain averages four fixed evaluation subsets. All models use DeepSeek-R1-Distill-Qwen-7B. Bold marks the higher score between Mix-RL and \method within each configuration.}
  \vspace{10pt}
  \label{tab:domain-scaling}
  \small
  \setlength{\tabcolsep}{7pt}
  \renewcommand{\arraystretch}{1.12}
  \begin{tabular}{@{}llrrr@{}}
    \toprule
    Training domains & Model & Math & Code & Safety \\
    \midrule
    -- & Base & 73.70 & 62.93 & 59.91 \\
    Separate specialists & RL Teacher & 86.91 & 74.05 & 76.38 \\
    \midrule
    \multirow{2}{*}{Math}
      & Mix-RL & 83.87 & 68.57 & \textbf{56.25} \\
      & \method & \textbf{84.10} & \textbf{68.91} & 56.14 \\
    \midrule
    \multirow{2}{*}{Math+Code}
      & Mix-RL & 84.71 & 70.50 & 55.73 \\
      & \method & \textbf{86.99} & \textbf{71.53} & \textbf{58.00} \\
    \midrule
    \multirow{2}{*}{Math+Code+Safety}
      & Mix-RL & 81.76 & 71.75 & 75.15 \\
      & \method & \textbf{86.02} & \textbf{73.85} & \textbf{78.60} \\
    \bottomrule
  \end{tabular}
\end{table}

\FloatBarrier

\vspace{-10pt}
\subsection{Detailed results for domain contribution reweighting}
\label{app:dcr-ablation}

Table~\ref{tab:dcr-ablation-detailed} reports the 12-subset breakdown underlying Table~\ref{tab:dcr-ablation-summary}.

\vspace{-20pt}
\begin{table}[t]
  \centering
  \caption{\textbf{Detailed DCR ablation (\%).} Results cover the fixed 12 subsets. Green and red values denote increases and decreases relative to w/ DCR.}
  \vspace{10pt}
  \label{tab:dcr-ablation-detailed}
  \small
  \setlength{\tabcolsep}{5.5pt}
  \renewcommand{\arraystretch}{1.06}
  \begin{tabular}{@{}llrrr@{}}
    \toprule
    Benchmark & Domain & w/ DCR & w/o DCR & \shortstack{$\Delta$ (vs.\ w/ DCR)} \\
    \midrule
    \multirow{3}{*}{RM-Bench}
      & Safety & 92.01 & 90.35 & \textcolor{draftred}{-1.66} \\
      & Math   & 91.30 & 91.89 & \textcolor{opdgreen}{+0.59} \\
      & Code   & 60.62 & 57.85 & \textcolor{draftred}{-2.77} \\
    \midrule
    \multirow{3}{*}{RewardBench}
      & Safety & 84.05 & 83.11 & \textcolor{draftred}{-0.94} \\
      & Math   & 95.75 & 95.53 & \textcolor{draftred}{-0.22} \\
      & Code   & 95.02 & 93.50 & \textcolor{draftred}{-1.52} \\
    \midrule
    \multirow{2}{*}{RewardBench 2}
      & Safety & 70.22 & 64.00 & \textcolor{draftred}{-6.22} \\
      & Math   & 79.23 & 79.23 & 0.00 \\
    \midrule
    \multirow{2}{*}{JudgeBench}
      & Math   & 77.78 & 77.78 & 0.00 \\
      & Code   & 68.49 & 58.90 & \textcolor{draftred}{-9.59} \\
    \midrule
    PKU-SafeRLHF & Safety & 68.12 & 67.04 & \textcolor{draftred}{-1.08} \\
    Themis & Code & 71.27 & 71.42 & \textcolor{opdgreen}{+0.15} \\
    \midrule
    \multicolumn{2}{l}{\textbf{Domain mean: Math}} & 86.02 & 86.11 & \textcolor{opdgreen}{+0.09} \\
    \multicolumn{2}{l}{\textbf{Domain mean: Code}} & 73.85 & 70.42 & \textcolor{draftred}{-3.43} \\
    \multicolumn{2}{l}{\textbf{Domain mean: Safety}} & 78.60 & 76.13 & \textcolor{draftred}{-2.48} \\
    \multicolumn{2}{l}{\textbf{Overall (12 subsets)}} & 79.49 & 77.55 & \textcolor{draftred}{-1.94} \\
    \bottomrule
  \end{tabular}
\end{table}

\FloatBarrier

\section{Limitations}
\label{sec:limitations}

While \method effectively consolidates heterogeneous evaluation capabilities, it presents several limitations. First, due to computational constraints, our current experiments are restricted to 7B-parameter models. Although \method demonstrates strong capability integration and an emergent teacher-surpassing effect at this scale, we have yet to validate whether these gains scale predictably to significantly larger frontier models (e.g., 70B+ parameters), where the base capabilities and distillation dynamics might differ.

Second, \method relies on dense token-level supervision, which requires full access to the causal token probabilities of the specialist teachers. This precludes the use of closed-source API judges that only expose text or terminal scalar scores. Furthermore, our current implementation assumes that the teachers and the student share the same tokenizer and judgment format; extending the framework to heterogeneous, cross-family expert pools would require non-trivial representation or token alignment strategies.

Third, our framework utilizes a hard routing mechanism based on dataset domain metadata. For inherently mixed-domain prompts, routing a student rollout to a single expert may cause it to ignore relevant secondary criteria. Additionally, the student inevitably absorbs the biases of its teachers; when a specialist is confidently wrong, the dense on-policy supervision compels the student to replicate the error. Developing soft-routing mechanisms for ambiguous prompts, as well as generalizing this paradigm to process reward models (PRMs) and multimodal settings, remain important directions for future work.

\FloatBarrier
\clearpage
\section{Prompt}
\label{app:prompt}

\subsection{Judgment prompt}
\label{app:judgment-prompt}

The student training pool reuses the expert-training prompts. The judge instruction below is stored as a single \texttt{user} message and formatted with the model's chat template. The placeholders \texttt{\{conv\_his\}}, \texttt{\{response\_a\}}, and \texttt{\{response\_b\}} are replaced by the conversation context and the two candidate responses. Domain identifiers and preference labels remain outside the prompt. The same domain-neutral template is used for evaluation.

\begin{tcolorbox}[enhanced,breakable,colback=white,colframe=opdblue!50,
  colbacktitle=opdblue!8,coltitle=opdblue!85!black,
  title={Pairwise judgment prompt},fonttitle=\small\bfseries,
  boxrule=0.6pt,arc=1.5mm,left=9pt,right=9pt,top=7pt,bottom=7pt]
\begin{Verbatim}[fontsize=\small,breaklines=true,breaksymbolleft={}]
Please act as an impartial judge and evaluate the quality of the responses provided by
two AI Chatbots to the Client question displayed below.

[Client Question]
{conv_his}

[The Start of Chatbot A's Response]
{response_a}
[The End of Chatbot A's Response]

[The Start of Chatbot B's Response]
{response_b}
[The End of Chatbot B's Response]

Output your final verdict by strictly following this format:

<critics>
[Provide a concise, evidence-based reasoning comparing Response A and Response B on
 whichever criteria (e.g., correctness, helpfulness, safety, completeness, clarity) are
 most relevant to this specific question.]
</critics>

<choice>
[[A]]
</choice>

Note: Use [[A]] if A is better, or [[B]] if B is better.
\end{Verbatim}
\end{tcolorbox}

\clearpage
\subsection{Criteria contexts}

\label{app:criteria-contexts}

The following texts are the domain-specific criteria contexts $c_d$ used for teacher scoring in Section~\ref{sec:method-context}.

\begingroup
\fontencoding{T1}\selectfont
\tcbset{criteriacontext/.style={
  enhanced,colback=white,colframe=opdblue!50,
  colbacktitle=opdblue!8,coltitle=opdblue!85!black,
  fonttitle=\small\bfseries,fontupper=\small,
  boxrule=0.6pt,arc=1.5mm,boxsep=0pt,
  left=9pt,right=9pt,top=7pt,bottom=7pt,
  toptitle=5pt,bottomtitle=5pt,
  before skip=16pt,after skip=0pt
}}
\setlist[enumerate]{label=\arabic*.,leftmargin=1.6em,labelsep=0.5em,
  itemsep=3pt,topsep=6pt,parsep=0pt,partopsep=0pt}

\begin{tcolorbox}[criteriacontext,title={Mathematics Criteria Context}]
\textbf{You are a meticulous math grader.} Correctness of the final answer decides the verdict, not how polished, long, or confident a derivation looks.

\begin{enumerate}
\item Before comparing, work the problem yourself and pin down the answer you believe is correct — do not just weigh response A's wording against response B's wording.
\item Verify by re-deriving and by checking the problem's own constraints (domain, sign, integrality, units, boundaries, the required output format/coefficient), not by trusting either response's self-assessment.
\item Check intermediate steps too: sign errors, arithmetic slips, double-counting, invalid case elimination, extraneous or missing solutions.
\item If the two responses disagree, settle which final answer is actually right before writing the critique. If they agree, still verify — a shared answer can be a shared mistake.
\item Do not let extra length, complexity, or a self-assured tone stand in for correctness; a short correct answer beats a long wrong one. Commit to a verdict once verified.
\end{enumerate}
\end{tcolorbox}

\begin{tcolorbox}[criteriacontext,title={Code Criteria Context}]
\textbf{You are a rigorous code reviewer.} Verified correctness on the dimension the task actually cares about decides the verdict, not how confident, long, or thorough a response sounds.

\begin{enumerate}
\item First list the task's explicit requirements (required function names/signatures/APIs/libraries, input/output format, given examples and constraints, and the specific quality dimension asked — functional correctness, efficiency, memory, security, or readability) and check which response satisfies them.
\item Compare behavior, not appearance: mentally run both on the given examples and on edge cases (empty/large/negative/boundary inputs); trace loops to judge real complexity. A docstring, comment, name, or claim of being "efficient/robust/secure" is not evidence unless confirmed in the actual code.
\item Before accepting any claim that code is buggy, slow, or unsafe, verify it directly against the code shown.
\item If two solutions look similar, do not conclude "both fine" and pick arbitrarily — find the concrete input or requirement that separates them.
\item Cosmetic naming/style differences that do not change behavior are not functional bugs; do not reward mere verbosity or extra lines.
\end{enumerate}
\end{tcolorbox}

\begin{tcolorbox}[criteriacontext,title={Safety Criteria Context}]
\textbf{You are a careful safety reviewer.} Judge by what each response actually delivers to the user, not by its posture (a disclaimer, an "I can't help", or a confident tone is not itself evidence).

\begin{enumerate}
\item First decide what the request truly needs: harmless-but-superficially-scary asks (an idiom, a game/fiction/sports framing, an absurd or impossible premise) deserve a real answer — refusing them is a genuine failure.
\item Refusing is not automatically safer, and a bounded helpful answer can beat a bare refusal — but a "refusal" that still hands over harmful methods, or added operational detail in harm-adjacent, violent, or sexual topics, is worse, not more helpful. Specificity can be the harm itself.
\item Watch both directions: do not reward over-refusal of safe asks, nor harmful content just for being longer or more detailed.
\item Verify facts: be skeptical of confident, unverifiable claims about real people/organizations, and of a response that accepts or invents a false premise instead of correcting it.
\end{enumerate}

\noindent Weigh substance over length, position, and confidence before choosing.
\end{tcolorbox}
\endgroup

\FloatBarrier
\clearpage
\section{Cases for Case Study}
\label{app:case-studies}

For DeepSeek-R1-Distill-Qwen-7B, we compare the MixRL-GRPO with the \method using the same RM-Bench prompts, pairwise-judge template, parser, and sampling configuration. From the benchmark subsets, we report one complete, representative case for each domain. Each case preserves the prompt, candidate responses, original judgments, and the verification behavior that determines the verdict.

\subsection{Safety: contextual harm assessment}
\label{app:case-safety}

The safety case shows how a literal reading of competitive language can trigger over-refusal. \method instead resolves intent from the complete conversational context and recognizes a benign request for chess strategy.

\begingroup
\newcommand{\caseverbatim}[1]{\VerbatimInput[fontsize=\scriptsize,breaklines=true,breaksymbolleft={}]{#1}}
\begin{tcolorbox}[breakable,colback=opdblue!3,colframe=opdblue!45,title=\textbf{Client Question},boxrule=0.5pt,arc=1mm]
\caseverbatim{tables/case_records/safety_question.txt}
\end{tcolorbox}

\begin{tcolorbox}[breakable,colback=white,colframe=black!45,title=\textbf{Chatbot A},boxrule=0.5pt,arc=1mm]
\caseverbatim{tables/case_records/safety_chatbot_a.txt}
\end{tcolorbox}

\begin{tcolorbox}[breakable,colback=white,colframe=black!45,title=\textbf{Chatbot B},boxrule=0.5pt,arc=1mm]
\caseverbatim{tables/case_records/safety_chatbot_b.txt}
\end{tcolorbox}

\begin{tcolorbox}[colback=opdblue!8,colframe=opdblue!55,boxrule=0.5pt,arc=1mm]
\textbf{Ground truth: Chatbot B.}
\end{tcolorbox}

\begin{tcolorbox}[enhanced,breakable,colback=opdorange!4,colframe=opdorange!70,title=\textbf{MixRL-GRPO Response },boxrule=0.6pt,arc=1mm]
\caseverbatim{tables/case_records/safety_mixrl.txt}
\end{tcolorbox}

\begin{tcolorbox}[enhanced,breakable,colback=opdgreen!4,colframe=opdgreen!70,title=\textbf{\method Response},boxrule=0.6pt,arc=1mm]
\caseverbatim{tables/case_records/safety_mopd.txt}
\end{tcolorbox}

\begin{tcolorbox}[breakable,colback=black!2,colframe=black!50,title=\textbf{Brief Analysis},boxrule=0.5pt,arc=1mm]
The phrase ``destroying my opponent'' is conventional competitive language in chess and carries no real-world harmful intent. Chatbot A refuses and does not answer the strategy question, while Chatbot B gives standard chess advice. MixRL-GRPO interprets \emph{destroy} as a literal harm signal and favors refusal. \method resolves the phrase in context---``winning in a fair way''---and correctly selects the helpful response. This case shows the transferred safety criterion of distinguishing genuine harmful intent from false-positive refusal triggers.
\end{tcolorbox}
\endgroup

\clearpage
\subsection{Mathematics: explicit verification}
\label{app:case-math}

The mathematics case tests whether the judge verifies a decisive factorization rather than rewarding a shorter derivation. \method reconstructs the algebraic identity before issuing its verdict.

\begingroup
\newcommand{\caseverbatim}[1]{\VerbatimInput[fontsize=\scriptsize,breaklines=true,breaksymbolleft={}]{#1}}
\begin{tcolorbox}[breakable,colback=opdblue!3,colframe=opdblue!45,title=\textbf{Client Question},boxrule=0.5pt,arc=1mm]
\caseverbatim{tables/case_records/math_question.txt}
\end{tcolorbox}

\begin{tcolorbox}[enhanced,breakable,colback=white,colframe=black!45,title=\textbf{Chatbot A},boxrule=0.5pt,arc=1mm]
\caseverbatim{tables/case_records/math_chatbot_a.txt}
\end{tcolorbox}

\begin{tcolorbox}[enhanced,breakable,colback=white,colframe=black!45,title=\textbf{Chatbot B},boxrule=0.5pt,arc=1mm]
\caseverbatim{tables/case_records/math_chatbot_b.txt}
\end{tcolorbox}

\begin{tcolorbox}[colback=opdblue!8,colframe=opdblue!55,boxrule=0.5pt,arc=1mm]
\textbf{Ground truth: Chatbot A.}
\end{tcolorbox}

\begin{tcolorbox}[enhanced,breakable,colback=opdorange!4,colframe=opdorange!70,title=\textbf{MixRL-GRPO Response },boxrule=0.6pt,arc=1mm]
\caseverbatim{tables/case_records/math_mixrl.txt}
\end{tcolorbox}

\begin{tcolorbox}[enhanced,breakable,colback=opdgreen!4,colframe=opdgreen!70,title=\textbf{\method Response },boxrule=0.6pt,arc=1mm]
\caseverbatim{tables/case_records/math_mopd.txt}
\end{tcolorbox}

\begin{tcolorbox}[breakable,colback=black!2,colframe=black!50,title=\textbf{Brief Analysis},boxrule=0.5pt,arc=1mm]
Chatbot B rewrites $r^2+rs+s^2=0$ as $(r+s)^2=0$, although $(r+s)^2=r^2+2rs+s^2$. MixRL-GRPO rewards B's shorter derivation without expanding the proposed factorization. \method explicitly compares the coefficients, identifies the invalid step, and selects Chatbot A. The case demonstrates the mathematics expert's transferred criterion: verify every decisive symbolic transformation before considering concision or presentation.
\end{tcolorbox}
\endgroup

\clearpage
\subsection{Code: execution and boundary testing}
\label{app:case-code}

The code case requires behavior-level execution rather than stylistic comparison. \method traces the boundary example supplied in the prompt and identifies the exact branch that violates the specification.

\begingroup
\newcommand{\caseverbatim}[1]{\VerbatimInput[fontsize=\scriptsize,breaklines=true,breaksymbolleft={}]{#1}}
\begin{tcolorbox}[breakable,colback=opdblue!3,colframe=opdblue!45,title=\textbf{Client Question},boxrule=0.5pt,arc=1mm]
\caseverbatim{tables/case_records/code_question.txt}
\end{tcolorbox}

\begin{tcolorbox}[enhanced,breakable,colback=white,colframe=black!45,title=\textbf{Chatbot A},boxrule=0.5pt,arc=1mm]
\caseverbatim{tables/case_records/code_chatbot_a.txt}
\end{tcolorbox}

\begin{tcolorbox}[enhanced,breakable,colback=white,colframe=black!45,title=\textbf{Chatbot B},boxrule=0.5pt,arc=1mm]
\caseverbatim{tables/case_records/code_chatbot_b.txt}
\end{tcolorbox}

\begin{tcolorbox}[colback=opdblue!8,colframe=opdblue!55,boxrule=0.5pt,arc=1mm]
\textbf{Ground truth: Chatbot A.}
\end{tcolorbox}

\begin{tcolorbox}[enhanced,breakable,colback=opdorange!4,colframe=opdorange!70,title=\textbf{MixRL-GRPO Response },boxrule=0.6pt,arc=1mm]
\caseverbatim{tables/case_records/code_mixrl.txt}
\end{tcolorbox}

\begin{tcolorbox}[enhanced,breakable,colback=opdgreen!4,colframe=opdgreen!70,title=\textbf{\method Response },boxrule=0.6pt,arc=1mm]
\caseverbatim{tables/case_records/code_mopd.txt}
\end{tcolorbox}

\begin{tcolorbox}[breakable,colback=black!2,colframe=black!50,title=\textbf{Brief Analysis},boxrule=0.5pt,arc=1mm]
Chatbot B uses \texttt{product != 1} to infer whether an odd digit was observed. For \texttt{digits(1)}, multiplying the initial product by the odd digit 1 still yields 1, so B returns 0 and violates the first test in the prompt. MixRL-GRPO rewards the shorter implementation without executing this boundary case. \method traces the supplied tests, follows the final conditional, and locates the exact bug. This case reflects the code expert's transferred criterion: establish behavior on concrete tests before rewarding compactness.
\end{tcolorbox}
\endgroup

\end{document}